\documentclass[letterpaper, 10 pt, conference]{ieeeconf}

\IEEEoverridecommandlockouts %
\usepackage{amsmath} %
\usepackage{amssymb} %
\usepackage{amsfonts}
\usepackage[usenames,dvipsnames]{xcolor}
\usepackage{graphicx}
\usepackage{tabularx}
\usepackage{multirow}
\usepackage{mathtools}
\usepackage{esvect} %
\usepackage[binary-units=true]{siunitx}
\usepackage{caption}
\usepackage[pdftex, pdfstartview={FitV}, pdfpagelayout={TwoColumnLeft},bookmarksopen=true,plainpages = false, colorlinks=true, linkcolor=black, citecolor = black, urlcolor = black,filecolor=black , pagebackref=false,hypertexnames=false, plainpages=false, pdfpagelabels ]{hyperref}
\usepackage[T1]{fontenc} %
\usepackage{mathtools, cuted} %

\usepackage{balance} %
\usepackage{booktabs} %
\usepackage[export]{adjustbox} %

\newcolumntype{x}[1]{%
>{\raggedleft\hspace{0pt}}p{#1}}%

\usepackage{algorithm}
\usepackage[noend]{algpseudocode} %
\definecolor{commentclr}{RGB}{34, 139, 34}
\newcommand{\algcomment}[1]{{\color{commentclr}\% #1}}

\usepackage{tikz}

\usepackage{subcaption}
\DeclareCaptionLabelSeparator{periodspace}{.\quad}
\usepackage{amsthm}

\theoremstyle{definition}

\usepackage{letltxmacro}
\LetLtxMacro\orgvdots\vdots
\LetLtxMacro\orgddots\ddots

\makeatletter
\DeclareRobustCommand\vdots{%
	\mathpalette\@vdots{}%
}
\newcommand*{\@vdots}[2]{%
	\sbox0{$#1\cdotp\cdotp\cdotp\m@th$}%
	\sbox2{$#1.\m@th$}%
	\vbox{%
		\dimen@=\wd0 %
		\advance\dimen@ -3\ht2 %
		\kern.5\dimen@
		\dimen@=\wd2 %
		\advance\dimen@ -\ht2 %
		\dimen2=\wd0 %
		\advance\dimen2 -\dimen@
		\vbox to \dimen2{%
			\offinterlineskip
			\copy2 \vfill\copy2 \vfill\copy2 %
		}%
	}%
}
\DeclareRobustCommand\ddots{%
	\mathinner{%
		\mathpalette\@ddots{}%
		\mkern\thinmuskip
	}%
}
\newcommand*{\@ddots}[2]{%
	\sbox0{$#1\cdotp\cdotp\cdotp\m@th$}%
	\sbox2{$#1.\m@th$}%
	\vbox{%
		\dimen@=\wd0 %
		\advance\dimen@ -3\ht2 %
		\kern.5\dimen@
		\dimen@=\wd2 %
		\advance\dimen@ -\ht2 %
		\dimen2=\wd0 %
		\advance\dimen2 -\dimen@
		\vbox to \dimen2{%
			\offinterlineskip
			\hbox{$#1\mathpunct{.}\m@th$}%
			\vfill
			\hbox{$#1\mathpunct{\kern\wd2}\mathpunct{.}\m@th$}%
			\vfill
			\hbox{$#1\mathpunct{\kern\wd2}\mathpunct{\kern\wd2}\mathpunct{.}\m@th$}%
		}%
	}%
}
\makeatother

\let\oldnl\nl%
\newcommand{\nonl}{\renewcommand{\nl}{\let\nl\oldnl}}%
\makeatother

\def\br{\mathbb R}

\def\sA{\mathcal{A}}

\def\sG{\mathcal{G}}

\def\sl{\mathcal{L}}

\newcommand\eqdef{\mathrel{\overset{\makebox[0pt]{\mbox{\normalfont\tiny def}}}{=}}}

\graphicspath{{figures/}}
\title{\LARGE \bf
CLIPPER: A Graph-Theoretic Framework for Robust Data Association
	}

\author{Parker C. Lusk*, Kaveh Fathian*, Jonathan P. How%
	\thanks{P.\ C.\ Lusk, K.\ Fathian and J.\ P.\ How are with the Department of Aeronautics and Astronautics, Massachusetts Institute of Technology.
	    {\{plusk, kavehf, jhow\}@mit.edu.} *Authors contributed equally.}
    \thanks{This work is supported by Ford Motor Company and by ARL DCIST under Cooperative
    Agreement Number W911NF-17-2-0181.}
}%

\begin{document}

\maketitle
\thispagestyle{plain}
\pagestyle{plain}

\begin{abstract}
We present CLIPPER (Consistent LInking, Pruning, and Pairwise Error Rectification), a framework for robust data association in the presence of noise and outliers. 
We formulate the problem in a graph-theoretic framework using the notion of geometric consistency.
State-of-the-art techniques that use this framework utilize either combinatorial optimization techniques that do not scale well to large-sized problems, or use heuristic approximations that yield low accuracy in high-noise, high-outlier regimes. 
In contrast, CLIPPER uses a relaxation of the combinatorial problem and returns solutions that are guaranteed to correspond to the optima of the original problem.
Low time complexity is achieved with an efficient projected gradient ascent approach.
Experiments indicate that CLIPPER maintains a consistently low runtime of 15 ms where exact methods can require up to 24 s at their peak, even on small-sized problems with 200 associations.
When evaluated on noisy point cloud registration problems, CLIPPER achieves 100\% precision and 98\% recall in 90\% outlier regimes while competing algorithms begin degrading by 70\% outliers.
In an instance of associating noisy points of the Stanford Bunny with 990 outlier associations and only 10 inlier associations, CLIPPER successfully returns \hbox{8 inlier} associations with 100\% precision in 138 ms.
Code is available at \href{https://mit-acl.github.io/clipper}{\color{blue}https://mit-acl.github.io/clipper}.
\end{abstract}

\section{INTRODUCTION}\label{sec:intro}

\begin{figure}[t!]
\centering%
	\begin{subfigure}[b]{0.49\columnwidth}
	    \includegraphics[width=0.48\textwidth]{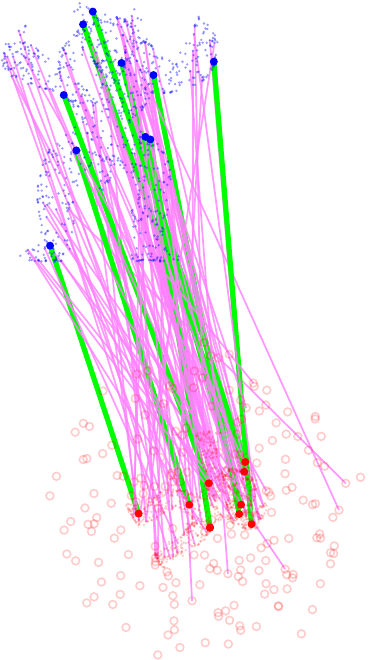}
	    \includegraphics[width=0.48\textwidth]{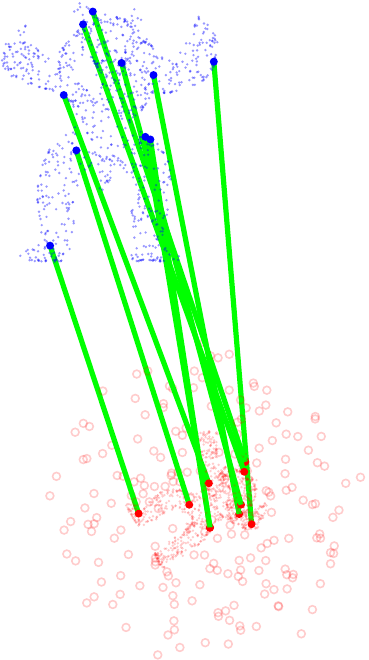}
	    \caption{Scaled point cloud registration}
	    \label{fig:applications:scalecloud}
	\end{subfigure}\hfill
	\begin{subfigure}[b]{0.44\columnwidth}
	    \centering
	    \includegraphics[width=0.85\textwidth]{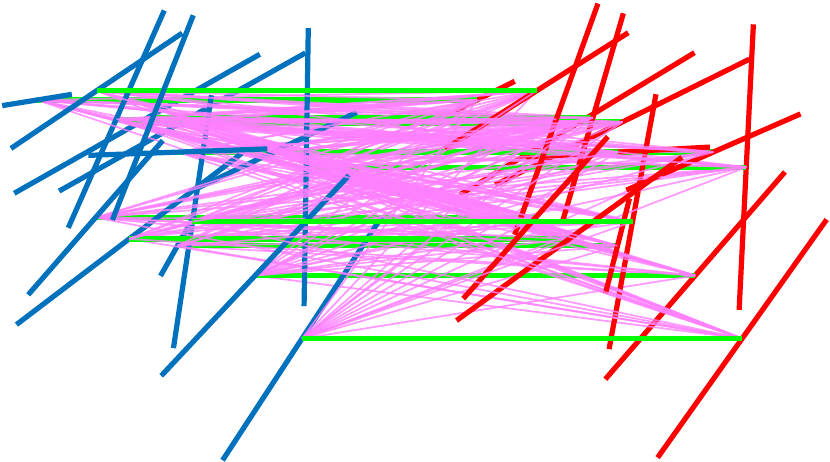}
	    \includegraphics[width=0.85\textwidth]{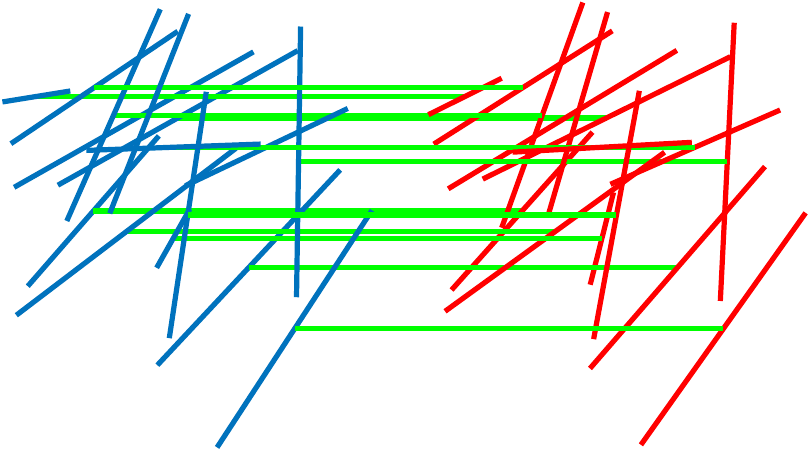}
	    \caption{Line cloud registration}
	    \label{fig:applications:linecloud}
	\end{subfigure}
	\begin{subfigure}[b]{0.44\columnwidth}
	    \centering
	    \includegraphics[width=0.6\textwidth,trim=5mm 5mm 10mm 10mm,clip,angle=90]{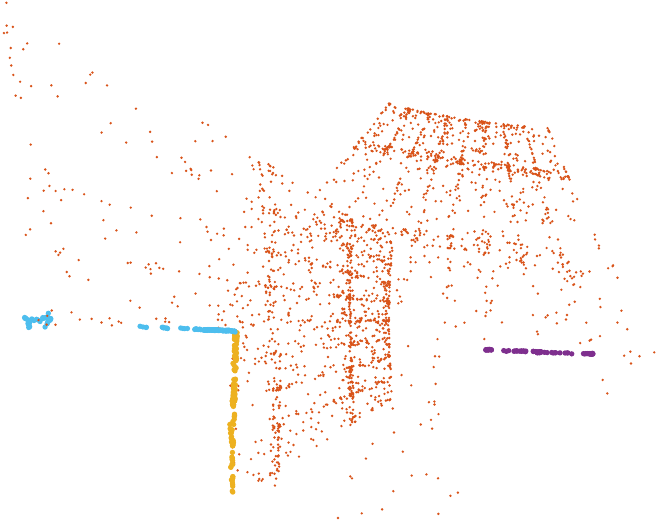}
	    \caption{Plane registration (top-down view)}
	    \label{fig:applications:planecloud}
	\end{subfigure}\hfill
	\begin{subfigure}[b]{0.50\columnwidth}
	    \includegraphics[width=0.43\textwidth,trim=12mm 20mm 12mm 22mm, clip,angle=0]{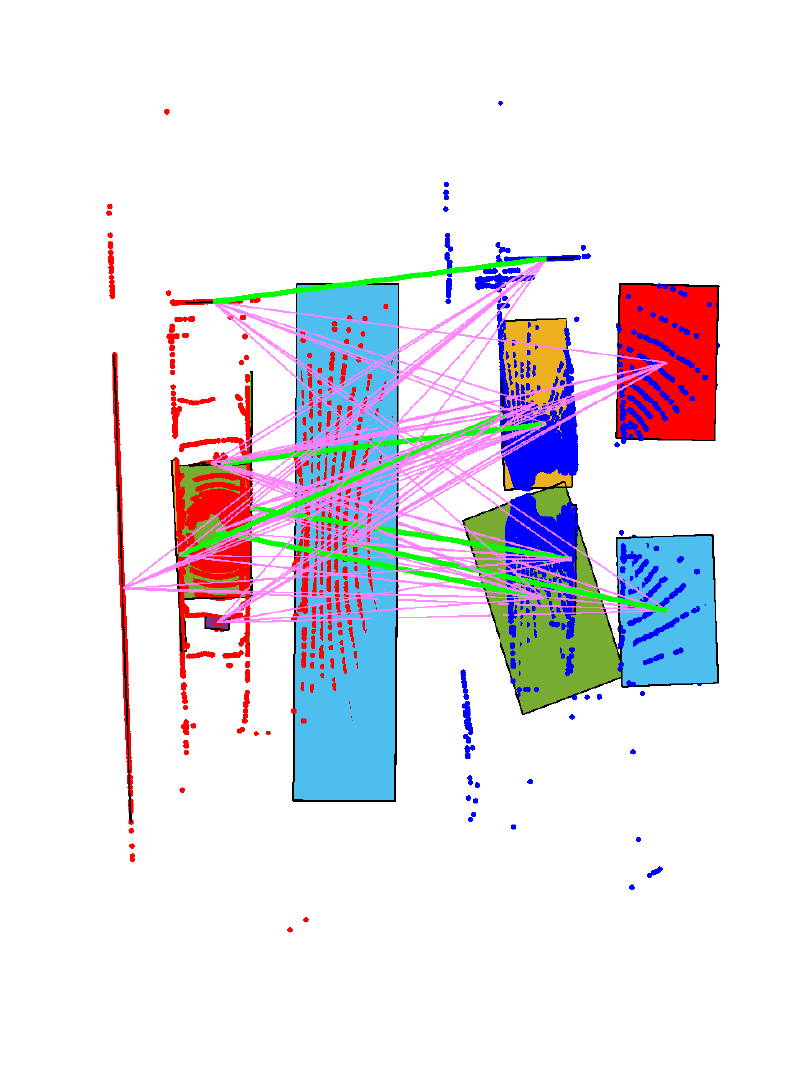}
	    \hfill
	    \includegraphics[width=0.43\textwidth,trim=12mm 20mm 12mm 22mm, clip,angle=0]{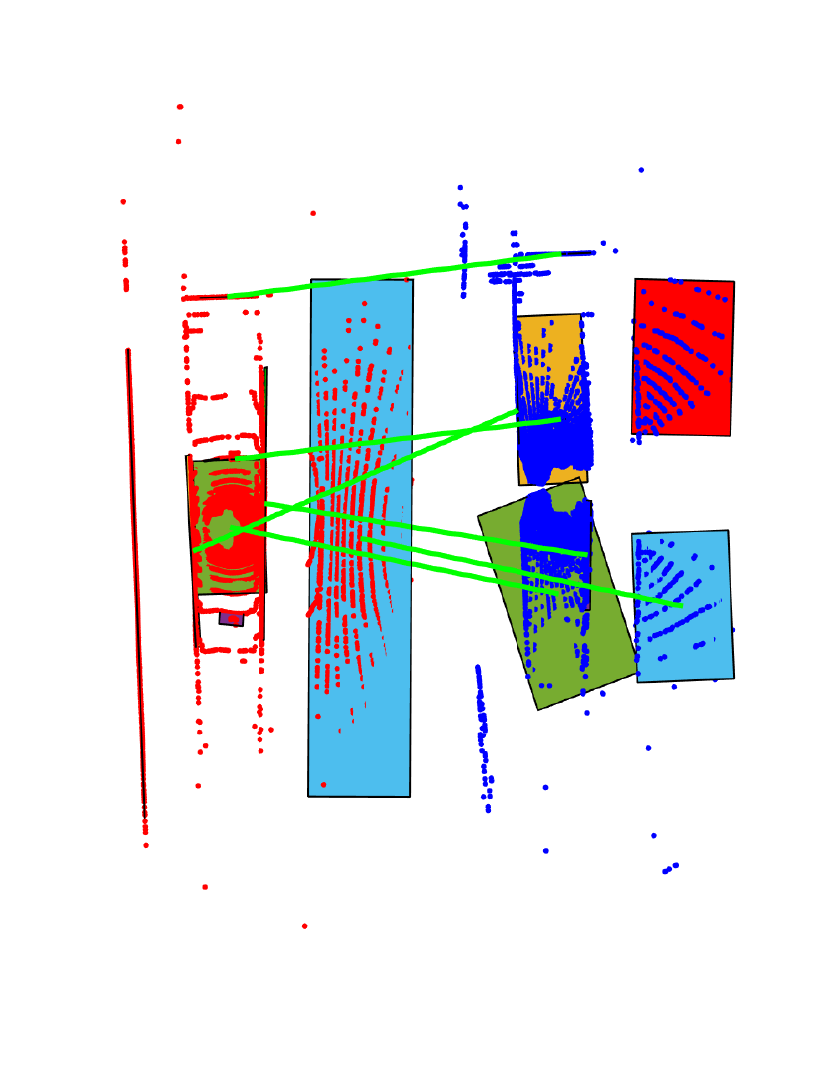}
	    \caption{Planar patch registration}
	    \label{fig:applications:patchcloud}
	\end{subfigure}
\caption{Applications were the CLIPPER framework is used for robust data association: (a) noisy point cloud registration with unknown scale and outliers; (b) noisy line cloud registration; (c) plane cloud registration for LiDAR sensor calibration in outdoor, urban environment (planes indicated by colored points, sensor scans are correctly registered); (d) planar patch registration (extracted from LiDAR scans of~\cite{agarwal2020ford}).
In (a), (b), (c) magenta and green lines indicate incorrect and correct associations, respectively.
Input associations can be generated from an external matching procedure or an all-to-all hypothesis in the case of no prior information.
Each setting generates a consistency graph that CLIPPER operates on to identify which of the input associations are the most geometrically consistent.
}
    \vspace*{-0.3em}
	\label{fig:applications}
\end{figure}

Finding correct one-to-one correspondences between two sets of objects $\sA \eqdef \{a_1, \dots, a_n\}$ and ${\sA' \eqdef \{a'_1, \dots, a'_m\}}$ is a fundamental problem in robotics, arising in a wide range of perception and estimation pipelines.
In practice, observations of objects are ``noisy'' and ``partial'', i.e., 
when an \textit{unknown} number of objects in $\sA$ do not correspond to any object in $\sA'$ (outliers).
The traditional linear assignment approach based on corresponding objects of high similarity, e.g., using the Hungarian~\cite{kuhn1955hungarian} or auction~\cite{bertsekas1988auction} algorithms, is not robust to high-noise, high-outlier regimes, leading to incorrect correspondences.
In this work, we propose the CLIPPER (Consistent LInking, Pruning, and Pairwise Error Rectification) framework which leverages the notion of \textit{geometric consistency} between object pairs to find correct correspondences in these extreme regimes. 
Our primary motivation is robust perception with applications such as those shown in Fig.~\ref{fig:applications}, SLAM/loop closure~\cite{lajoie2020door,mangelson2018pairwise}, point cloud registration~\cite{bustos2017gore,yang2020teaser}, shape alignment~\cite{li2011robustly}, object detection~\cite{qi2017pointnet}, and multiple object tracking~\cite{chiu2020auto3dmot,kim2015mht}.
However, our framework is general and applicable to other pairwise data association problems.

\begin{figure*}[th!]
	\centering
	\begin{subfigure}[b]{0.99\textwidth}\includegraphics[trim = 0mm 0mm 0mm 0mm, clip, width=1\textwidth] {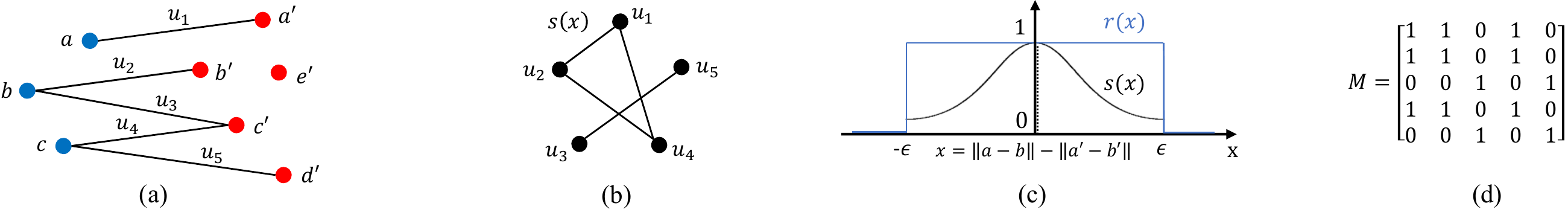}
	\end{subfigure}%
	\caption{An illustrative example of the consistency graph and its affinity matrix for point cloud registration. 
	(a) Blue and red point clouds where putative associations are labeled as $u_1, \dots, u_5$. 
	(b) The consistency graph $\mathcal{G}$ with vertices representing the associations and edges between two vertices indicating their geometric consistency. In the noiseless case, any two associations $u_1,u_2$ mapping points $a,b$ to $a',b'$ are consistent if $\|a \,\text{-}\, b \| = \| a' \,\text{-}\, b'\|$.
	(c) Edges of the consistency graph are weighted to measure the consistency of  associations using scalar functions $s(x)$ or $r(x)$. For example, the edge connecting vertices $u_1,u_2$ takes the weight $s(\|a \,\text{-}\, b \| \,\text{-}\, \| a' \,\text{-}\, b'\|)$. When the difference between distances is larger than the threshold $\epsilon$ or when two associations start and end at the same point, they are deemed as inconsistent. 
	 (d) The affinity matrix $M$ of the consistency graph $\mathcal{G}$, computed for the point cloud example using the binary score $r(x)$. }
	\vspace*{-1.5em}
	\label{fig:consistency_graph}	
\end{figure*}

When an attribute between objects in set $\sA$ is the same as the attribute between their correctly associated objects in $\sA'$, these objects are considered geometrically consistent (e.g., see Fig.~\ref{fig:consistency_graph}).
Incorporating geometric consistency in data association ultimately leads to a combinatorial optimization, such as maximum clique~\cite{bailey2000data}, maximum consensus~\cite{antonante2020outlier,wen2020simultaneous}, or quadratic assignment~\cite{bernard2019synchronisation} formulations.
Relaxations of this NP-hard problem exist~\cite{leordeanu2005spectral,le2019sdrsac}, but exhibit poor performance in high-outlier regimes or for large problem size.
In contrast, CLIPPER maintains high precision with low runtime across various outlier-regimes and problem sizes.

CLIPPER incorporates the concept of geometric consistency in a graph-theoretic framework, and proposes finding consistent association (inliers) by finding the densest subgraph.
This formulation is particularly suited to weighted graphs and improves precision compared to maximum clique frameworks which are limited to binary graphs.
CLIPPER uses the continuous relaxation technique introduced in~\cite{belachew2017nmfmcp} to guarantee that the recovered solution corresponds to a dense subgraph.
Using projected gradient ascent with backtracking line search, CLIPPER achieves a consistently low runtime as compared to other algorithms.
Further, our experiments show that CLIPPER is capable of attaining 100\% precision even in high outlier regimes where other algorithms breakdown.

In summary, the contributions of this work include:
\begin{itemize}
    \item An optimization formulation for selecting inlier associations suitable for both binary and weighted graphs.
    \item A relaxation of the resulting NP-hard optimization problem with optimality guarantees (see Sec~\ref{sec:continuous-relaxation}).
    \item A polynomial-time algorithm for solving the relaxed formulation based on projected gradient ascent, scalable to large-sized data association problems.
\end{itemize}
Additionally, we include a discussion on the application of geometric consistency to various observation types (e.g., points, lines, planes) commonly found in robotic perception.
Finally, we benchmark CLIPPER against the state of the art when finding associations between two point clouds.
We find that CLIPPER is able to achieve 100\% precision in 99\% outlier regimes, where the performance of competing algorithms begin to degrade.

Related works most pertinent to this work include Bailey et al.~\cite{bailey2000data}, where geometric consistency is leveraged for 2D LiDAR scan matching.
Formulated as a maximum common subgraph problem, their technique resulted in a \textit{binary} consistency graph for which the maximum clique indicated the correct data association.
Leordeanu and Hebert~\cite{leordeanu2005spectral} built on this graph-theoretic idea and instead constructed a \textit{weighted} consistency graph, where edge weights represent the geometric consistency of associations.
Enqvist et al.~\cite{enqvist2009optimal} noted the suboptimality of~\cite{leordeanu2005spectral} and proposed a vertex covering formulation, essentially an alternative to the maximum clique formulation of~\cite{bailey2000data}.
Parra et al.~\cite{bustos2019practical} proposed a practical maximum clique algorithm for geometric consistency based on branch and bound and graph coloring.
Recent algorithms leveraging these ideas for estimation tasks include PCM~\cite{mangelson2018pairwise}, where maximum cliques correspond to the largest set of pairwise-consistent loop closure measurements, and TEASER~\cite{yang2020teaser}, where maximum cliques correspond to inlier associations for point cloud registration, further formalized in ROBIN~\cite{shi2020robin}.
CLIPPER advances these works with a continuous relaxation applicable to weighted graphs.

\section{Graph-Theoretic Formulation}\label{sec:probform}

A standard approach for robust data association in the presence of noise and outliers is to find the largest set of \textit{geometrically consistent} associations. 
This problem can be formulated in a graph-theoretic framework.
In this section, we review the consistency graph framework and the construction of the affinity matrix.  

\subsection{Consistency Graph}\label{sec:consistency-graph}

For simplicity, and without loss of generality, we introduce the concept of a consistency graph for the point cloud registration problem.
In Section~\ref{sec:applications}, we will show how this general framework can be used for robust data association in other robotic perception domains. 
The point cloud registration problem consists of finding the rotation and translation that map coordinates of a set of points to their corresponding points in another set.
The main challenge is finding the correct correspondences as the points are noisy and contain outliers.
\textit{Outlier points} are points with large noise, and ones that are partially observed (i.e., points in one set that do no have a counterpart in the other set).  
\textit{Outlier associations} are correspondences that match (inlier or outlier) points in one set incorrectly to points in the other set.  
Fig.~\ref{fig:consistency_graph}a gives an example of two point clouds identified by blue and red colors.  
Inlier points are denoted by $a,b,c$ in the blue point cloud and their transformed counterparts by $a', b', c'$ in the red point cloud.
Points $d', e'$ do not correspond to any blue point and hence are considered as outliers.
A set of putative associations between the points, denoted by $u_1, \dots, u_5$, are given, where $u_1, u_2, u_4$ are inlier and $u_3, u_5$ are outlier associations.
In the case that no putative associations are given, an all-to-all hypothesis can be generated.

The consistency of associations can be assessed and represented in the graph-theoretic framework of the \textit{consistency graph}. The consistency graph $\sG$ of $n$ associations consists of $n$ vertices, where each vertex represents an association. 
Edges between the vertices of $\mathcal{G}$ show that their corresponding associations are consistent. 
The example in Fig.~\ref{fig:consistency_graph}b illustrates the consistency graph for the associations in Fig.~\ref{fig:consistency_graph}a.
Since rotation and translation are distance-preserving transformations, 
the distance between the points in one set should be identical (in the noiseless setting) to their counterparts in the other set when associations are correct. 
This attribute can be used to assess the \textit{geometric consistency} of two associations, where an edge between two vertices of $\sG$ indicates that the distances between the points matched by the associations are the same. 
The largest set of mutually consistent associations is given by the largest fully connected subgraph (maximum clique), which consists of vertices $u_1,u_2,u_4$ in the example of Fig.~\ref{fig:consistency_graph}b.

\subsection{Affinity Matrix}\label{sec:affinity-matrix}

The \textit{affinity matrix} $M$ of a consistency graph with $n$ vertices is an $n\times n$ \textit{symmetric} matrix with entries in the interval $[0,1]$. 
The diagonal entries $M(i,i)$ measure the \textit{similarity} of the data points that association $i$ matches, e.g., based on the similarity of point ``descriptors'' (for instance, the FPFH descriptors for point clouds~\cite{rusu2009fast}).
Scores of $0$ and $1$ indicate the lowest and the highest similarity, respectively. The diagonal entries of $M$ are set to $1$ when similarity information is not available.
In this case, $M = A + I$, where $A$ is the (weighted) \textit{adjacency matrix} of the consistency graph and $I$ is the identity matrix. 
The off-diagonal entries $M(i,j)$ measure the geometric consistency of association pairs $i$ and $j$, and similarly range from $0$ to $1$. 
For example, in the point cloud registration problem, the distance between two points $a, b$ matched to $a', b'$  by associations $u_i$ and $u_j$ can be used to measure the consistency as 
$M(i,j) \eqdef s\left(\| a \,\text{-}\, b \| \,\text{-}\, \| a' \,\text{-}\, b'\| \right) \in [0,1]$. Here, $s: \br \rightarrow [0,1]$ is a scoring function such that $s(0) = 1$ and $s(x) = 0$ for $|x| > \epsilon$, as illustrated in Fig~\ref{fig:consistency_graph}c.
The threshold $\epsilon$ is based on a bounded noise model with a noise range of $\epsilon/2$ on the point coordinates.
Hence, if the distance between the points differs more than this threshold, the associations are considered inconsistent. 
Another source of inconsistency, referred to as the \textit{distinctness constraint}, is when correct associations are expected to be one-to-one. Hence, for any two associations $i$ and $j$ that originate or terminate at the same point, $M(i,j) \eqdef 0$ to indicate mutual inconsistency of associations.
Lastly, we note that a binary scoring function, e.g., $r(x): \br \rightarrow \{0,1\}$ in Fig.~\ref{fig:consistency_graph}c, can be used. This will lead to a binary affinity matrix, shown in Fig.~\ref{fig:consistency_graph}d.

\section{Optimization Formulation of CLIPPER}\label{sec:}

Given the consistency graph $\mathcal{G}$ with $n$ vertices representing associations (all-to-all or putative), and its $n\times n$, symmetric affinity matrix $M$, we propose the problem 
\begin{gather} \label{eq:mainProblem}
\begin{array}{ll}
\underset{u \in \{0,1\}^n}{\text{maximize}} & \dfrac{u^\top  M \, u}{u^\top u}
\\
\text{subject to} & u_i \, u_j = 0  \quad \text{if}~ M(i,j)=0, ~ \forall_{i,j},
\end{array}
\end{gather}
for finding the \textit{densest} subset of consistent associations.
Here, the optimization variable $u$ is a binary vector of size $n$, with $1$ elements indicating associations that are selected as inliers, and $0$'s otherwise.
Since $u$ is binary, the constraint $u_i \, u_j = 0$ ensures that if $M(i,j) = 0$, then at most one of the associations $u_i$ or $u_j$ is selected in the answer.

When $M$ is binary (e.g., obtained by using the scoring function $r(x)$ in Fig.\,\ref{fig:consistency_graph}c) and has one diagonal entries, it is straightforward to show that \eqref{eq:mainProblem} simplifies to 
\begin{gather} \label{eq:MCP}
\begin{array}{ll}
\underset{u \in \{0,1\}^n}{\text{maximize}} & \sum_{i = 1}^{n}{u_i}
\\
\text{subject to} & u_i \, u_j = 0  \quad \text{if}~ M(i,j)=0, ~ \forall_{i,j}.
\end{array}
\end{gather}
Problem~\eqref{eq:MCP} is known as the maximum clique problem (MCP), and its solution gives the largest set of consistent associations.  
The MCP framework is used frequently in the literature~\cite{yang2020teaser,mangelson2018pairwise,bailey2000data} for data association in the presence of noise and a large number of outliers, where the vertices/associations in the maximum clique are considered as the correct/inlier correspondences. 
The justification for this consideration is based on the assumption that the noise and outlier points are random, unbiased, and unstructured, thus they are not expected to form a large clique in the consistency graph~\cite{leordeanu2005spectral}.
We note that MCP is a well-known NP-hard problem \cite{wu2015review} in its full generality, hence, algorithms that rely on solving \eqref{eq:MCP} become computationally intractable as the problem size grows.

The \textit{density} of a graph is defined as 
the total sum of edge weights divided by the number of vertices.
The \textit{densest subgraph} is the subset of graph vertices and their corresponding edges that have the highest density. 
Given a graph $\sG'$ with affinity matrix $M'$ (with 1 diagonal entries), the densest subgraph of $\sG'$ is found from
\begin{gather} \label{eq:denseSubgraph}
\begin{array}{ll}
\underset{u \in \{0,1\}^n}{\text{maximize}} & \dfrac{u^\top  M' \, u}{u^\top u},
\end{array}
\end{gather}
where elements $u_i = 1$ in the solution correspond to the vertices in the densest subgraph~\cite{lee2010survey}.
Noting the similarity of the objectives in \eqref{eq:denseSubgraph} and 
\eqref{eq:mainProblem},
problem \eqref{eq:mainProblem} can be interpreted as finding the densest, fully connected subgraph of $\sG$.  
The full connectivity requirement is due to the constraints in \eqref{eq:mainProblem}, which prohibit the selection of vertices $u_i, u_j$ that are not connected, i.e., $M(i,j) = 0$.
The densest-subgraph objective in \eqref{eq:mainProblem} is crucial in the weighted case, and sets it apart from the maximum edge weighted clique problem~\cite{Hosseinian2017}.
For example, consider a weighted matrix $M$ and two solution candidates $u, \bar{u}$ as
\begin{equation} \label{eq:Mexample}
M =
\begin{bmatrix}
1 & 1 & 0 & 0 & 0 \\
1 & 1 & 0 & 0 & 0 \\
0 & 0 & 1 & 0.2 & 0.2 \\
0 & 0 & 0.2 & 1 & 0.2 \\
0 & 0 & 0.2 & 0.2 & 1 \\
\end{bmatrix}, 
~
u = 
\begin{bmatrix}
1 \\
1 \\
0 \\
0 \\
0 \\
\end{bmatrix},
~
\bar{u} = 
\begin{bmatrix}
0 \\
0 \\
1 \\
1 \\
1 \\
\end{bmatrix}.
\end{equation}
The MCP objective in \eqref{eq:MCP}, or the unnormalized objective of $u^\top M\, u$, returns $\bar{u}$ as the optimum solution, whereas the block of $M$ corresponding to $\bar{u}$ has low consistency scores of $0.2$ between the vertices/associations. 
On the other hand, the normalized objective of \eqref{eq:mainProblem} takes values of $2$ and $1.4$ for $u$ and $\bar{u}$, respectively, leading to selection of the smaller, but more consistent subgraph. 
Example~\eqref{eq:Mexample} further highlights the importance of choosing a weighted scheme over binary in the affinity matrix, 
since the maximum clique formulation returns $\bar{u}$ as a solution. 
While this problem can be avoided by choosing a smaller $\epsilon$, a conservative threshold leads to classifying correct associations as outliers (i.e., lower output recall). 
Further, a weighted scheme can resolve symmetric cases where two binary cliques have the same size.

\subsection{Continuous Relaxation}\label{sec:continuous-relaxation}

The main challenges in solving \eqref{eq:mainProblem} are the combinatorial complexity of the problem due to its binary domain and the nonlinearity of the objective in $u$. 
This makes it intractable to solve the problem to global optimality in real time, even for small-sized data association problems. 
A standard workaround is to relax the domain and the constraints of \eqref{eq:mainProblem} to obtain a continuous problem amenable to fast solution, followed by projecting this solution back to the domain and constraint manifold of the original problem.
Among the relaxation strategies that can be considered, the main advantage of the following approach is that the solutions obtained from the relaxed problem correspond to optima of the original problem, as will be discussed shortly.

We propose the relaxation of \eqref{eq:mainProblem} as  
\begin{gather} \label{eq:clipperProblem}
\begin{array}{ll}
\underset{u \in \br^n_+}{\text{maximize}} & F(u) \eqdef u^\top  M_d \, u 
\\
\text{subject to} & \| u \| \leq 1
\end{array}
\end{gather}
where $\br_+$ is the set of non-negative reals, $\| \cdot \|$ is the $\ell_2$ vector norm, and 
\begin{gather} \label{eq:Md}
M_d(i,j) \eqdef \left\{
\begin{array}{ll}
M(i,j) & \text{if} ~~ M(i,j) \neq 0 \\
-d & \text{if}  ~~ M(i,j) = 0
\end{array}
\right.
\end{gather}
where $d > 0$ is a positive scalar.
This approach is inspired by~\cite{belachew2017nmfmcp}, which directly integrates the constraints of the original problem into the continuous formulation of \eqref{eq:clipperProblem} via the matrix $M_d$. 
Intuitively, when $M_d(i,j) = -d$, the scalar $d$ penalizes joint selection of $u_i, u_j$ in the objective by the amount $-2 \,d \, u_i \, u_j$.
Hence, as $d$ increases the entries of solution $u$ that violate the constraints are pushed to zero. 

When $d \geq n$, (local or global) optima of \eqref{eq:clipperProblem} satisfy the constraints in the original problem, i.e., $u_i\, u_j = 0$ if $M(i, j) = 0$. 
This fact has been shown in~\cite{belachew2017nmfmcp} for the case when $M$ is a binary matrix.
While we have extended the proof to the weighted case, this discussion is beyond the space constraints of this paper and will be presented in a subsequent journal submission.
We remark that since \eqref{eq:mainProblem} is an NP-hard problem, depending on the initial condition an optimization algorithm used for solving \eqref{eq:clipperProblem} may converge to a local optima.
To guarantee finding the global optima, it would be required to search the entire space of solutions.

Given a solution $u$ of \eqref{eq:clipperProblem} with $d \geq n$, 
let $\sG' \subseteq \sG$ represent the subgraph that corresponds to the nonzero elements of $u$, and $M'$ be the submatrix of $M$ that corresponds to $\sG'$. 
Noting that $u$, and therefore $\sG'$, satisfies the constraints in \eqref{eq:mainProblem}, 
problem~\eqref{eq:mainProblem} reduces to binarizing $u$ such that the objective is maximized. 
This is equivalent to the densest subgraph problem of \eqref{eq:denseSubgraph}, where the goal is to find $\sG'' \subseteq \sG'$ that has the maximum density. 
The densest subgraph problem can be solved in polynomial time using existing algorithms \cite{goldberg1984finding},
but a good approximate answer can be obtained immediately by selecting the $\hat{\omega} \eqdef \mathrm{round}( u^\top M\, u)$ largest elements of $u$ as vertices of $\sG''$.
The justification follows from the well-known facts that $u^\top M\,u$, which is the spectral radius of $\sG'$, is a tight upper bound for the graph's density \cite{cvetkovic1980spectra} 
and nonzero elements of $u$, which form the principal eigenvector of $M'$, represent centrality of their corresponding vertices,
which is a measure of connectivity for a vertex in the graph~\cite{canright2004roles}.

\section{CLIPPER Algorithm}\label{sec:clipper}

The CLIPPER algorithm consists of 1) obtaining a solution $u$ of \eqref{eq:clipperProblem} via a projected gradient ascent approach with backtracking line search; and 2) estimating the densest cluster in $u$ by selecting the $\hat{\omega}$ largest elements.

Algorithm~\ref{alg:clipper} seeks a feasible subgraph by incrementally increasing the penalty parameter $d$ (Line 13) and solving \eqref{eq:clipperProblem} via gradient ascent (Lines 7-12).
Noting that any optimal solution $u$ lies on the boundary of $\|u\|\le1$, 
the constraint manifold can be reduced to $\mathbb{R}^n_+\cap S^n$, where $S^n$ is the unit sphere.
Instead of moving directly along the gradient $\nabla F(u) \eqdef 2 M_d \, u$, we first project onto the tangent space of $S^n$ at $u$ (Line 9) and move according to this orthogonal projection $\nabla F_\perp(u)$.
To move quickly in the search space, the step size $\alpha$ is chosen greedily so that if there is a $u_i$ to be penalized (i.e., $\nabla F_\perp<0$ and $u>0$), a gradient step would cause $u$ to hit the boundary of the positive orthant (Line 10).
If no such $u_i$ exists, then $\alpha$ is set such that the gradient update step gracefully degrades into a power iteration.
In either case, if $\alpha$ is chosen too large, backtracking line search is used to find an appropriate step (Line 11).
The solution is projected back onto the constraint manifold (Line 12) and gradient ascent continues. %
Once $u$ has converged for the current value of $M_d$, $d$ is incrementally increased (Line 13) and gradient ascent runs again with new objective $F(u)=u^\top M_d u$.
This process continues until $d \geq n$ and $u$ has converged. 
The convergence of the algorithm is guaranteed by the convergence property of the projected gradient approach~\cite{bertsekas1997nonlinear}.

The final step of CLIPPER selects the densest component of the subgraph $\sG'$ (Lines 14-15),
as explained in Section~\ref{sec:continuous-relaxation}.
Since for all $M_d(i,j) = -d$ elements $u_i,u_j$ in the solution $u$ satisfy $u_i\, u_j = 0$, then $u^\top M \, u = u^\top M_d \, u$.
The vertices of the densest component are then identified as the largest $\hat{\omega}$ elements of $u$.

\begin{algorithm}[t]
\caption{CLIPPER}
\label{alg:clipper}
\small
\begin{algorithmic}[1]
\State \textbf{Input} affinity matrix $M\in[0,1]^{n\times n}$ of $\mathcal{G}$
\State \textbf{Output} $\mathcal{G}''$, densest component of feasible subgraph $\mathcal{G}'\subseteq\mathcal{G}$
\State $u\gets\mathrm{rand}(n,1)$ \algcomment{initialize with uniform random in $[0,1]$}
\State $d\gets d_0$ \algcomment{initialize d small}
\While {$d$ not large enough}
    \State calculate $M_d$ via eq.~\eqref{eq:Md}
    \While {$u$ not converged}
        \State $\nabla F(u) = 2M_du$
        \State $\nabla F_\perp(u) = (I-uu^\top) \nabla F(u)$ \algcomment{orthogonal projection}
        \State $\alpha = \min \{ \alpha_i \eqdef |u_i / (\nabla F_\perp)_i| : (\nabla F_\perp)_i < 0 , u_i >0 \}$
        \State $u\gets u + \alpha\nabla F_\perp(u)$\ \algcomment{$\alpha$ via backtracking line search}
        \State $u\gets\max(u/\|u\|,0)$ \algcomment{project back onto $\mathbb{R}^n_+\cap S^n$}
    \EndWhile
    \State $d\gets d + \Delta d$ \algcomment{incrementally increase d}
\EndWhile
\State $\hat{\omega}\gets\mathrm{round}(u^\top M_d u)$ \algcomment{estimate cluster size using max eig}
\State $\mathcal{G}''\gets$ vertices corresponding to largest $\hat{\omega}$ elements of $u$

\end{algorithmic}
\end{algorithm}

\subsection{Computational Considerations}\label{sec:computational-considerations}
Neglecting constant factors, the computational cost of Algorithm~\ref{alg:clipper} is $\mathcal{O}(|E|)$ per iteration, where $|E|$ is the number of edges in the graph $\sG$.
Note that the orthogonal projection in Line 9 should be implemented as $\nabla F(u) - u\langle u, \nabla F(u)\rangle$, which costs $\mathcal{O}(|V|)$ operations, where $|V|$ is the number of vertices in the graph $\sG$.
Most of the time is spent computing the matrix-vector product $\nabla F(u)=M_du$.
All other operations run in at most $\mathcal{O}(|V|)$ operations.

\section{Constructing The Consistency Graph for Common Robotics Applications}\label{sec:applications}

CLIPPER can be applied to a broad array of data association problems found in robotics.
All that is required is to identify a \textit{geometric invariant} in the data, i.e., a quantity that is invariant under transformation.
This invariant feature is then used to score geometric consistency for which a consistency graph $\sG$ can be constructed and CLIPPER can be used to quickly remove outlier associations.
We briefly review how to score geometric consistency for the application examples in Fig.~\ref{fig:applications}.

\textbf{Point Clouds}
In Section~\ref{sec:consistency-graph} we described how two points seen in each cloud will have the same pairwise distance if the association is correct.
This idea can be extended to scaled point clouds by using three point correspondences to form triangles for which correct associations will preserve the ratio of side lengths.
This leads to a tensor formulation which can be marginalized into an $n\times n$ affinity matrix~\cite{park2013fast}.

\textbf{Line Clouds}
A line $l:(p,v)$ is given by a point $p\in\mathbb{R}^3$ and direction $v\in\mathbb{R}^3$.
Given two sets of lines $\{l_1,\dots,l_n\}$ and $\{l_1',\dots,l_m'\}$, a simple invariant feature is the angle between pairs of lines, resulting in a geometric consistency score defined by $s(|\mathrm{acos}\,v_i^\top v_j - \mathrm{acos}\,v_i'^\top v_j'|)$.
The point $p$ could also be incorporated to improve precision.

\textbf{Plane Clouds}
A plane $\pi:(n,d)$ is given by its normal $n\in\mathbb{R}^3$ and distance from the origin $d$.
An invariant feature of four planes is the four-way intersection point.
However, the requirement of choosing four plane correspondences increases computational complexity.
Instead, the simpler invariant of angle between normals $n_i,n_j$ can be used resulting in the same consistency score as for line clouds.

\textbf{Patch Clouds}
A cloud of planar patches, e.g., extracted from LiDAR using~\cite{araujo2020rspd}, additionally provides the centroid and area of each patch.
Although neither the centroid nor area are guaranteed to be invariant across views (e.g., partial view), these values can be used to assign a similarity score to corresponding planar patches by weighting the diagonal entries of the affinity matrix $M$.
Geometric consistency is scored based on pairs of normals as with plane clouds.

\section{EXPERIMENTS}\label{sec:experiments}

We first present precision and runtime comparisons of CLIPPER as an approximate maximum clique solver on binary graphs.
Then, we demonstrate CLIPPER's ability to perform data association in different outlier regimes and its strengths as a dense subgraph solver against the state of the art.
Comparisons are performed against Leordeanu \& Hebert~\cite{leordeanu2005spectral} and Belachew \& Gillis~\cite{belachew2017nmfmcp}, which can also operate on weighted graphs.
The exact parallel maximum clique (PMCx) solver~\cite{rossi2015parallel} and its initial heuristic (PMCh) step are also compared against, as it has recently been used for data association in~\cite{yang2020teaser}.
We include results for CLIPPER when given a binary graph (CLIPPERb).
Experiments are run in MATLAB on an i9-7920, 64 GB RAM with a C++ interface to PMC.
We allowed 12 threads for PMC.

\begin{figure}[t!]
\centering
\includegraphics[width=1\columnwidth]{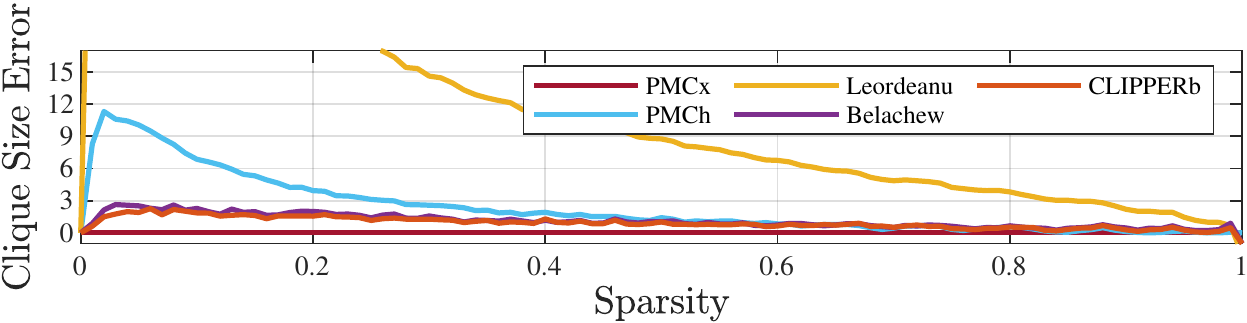}
\caption{
    Average clique size error of algorithms on binary graphs of size $n=200$ across $50$ Monte Carlo trials.
    PMC Exact recovers the true maximum clique at the cost of runtime.
    Leordeanu peaks in error with an underestimate of clique size by $57$.
    CLIPPER and Belachew consistently achieve the lowest error.
}
\label{fig:werr_vs_sparsity}
\end{figure}

\begin{figure}[t!]
\centering
\includegraphics[width=1\columnwidth]{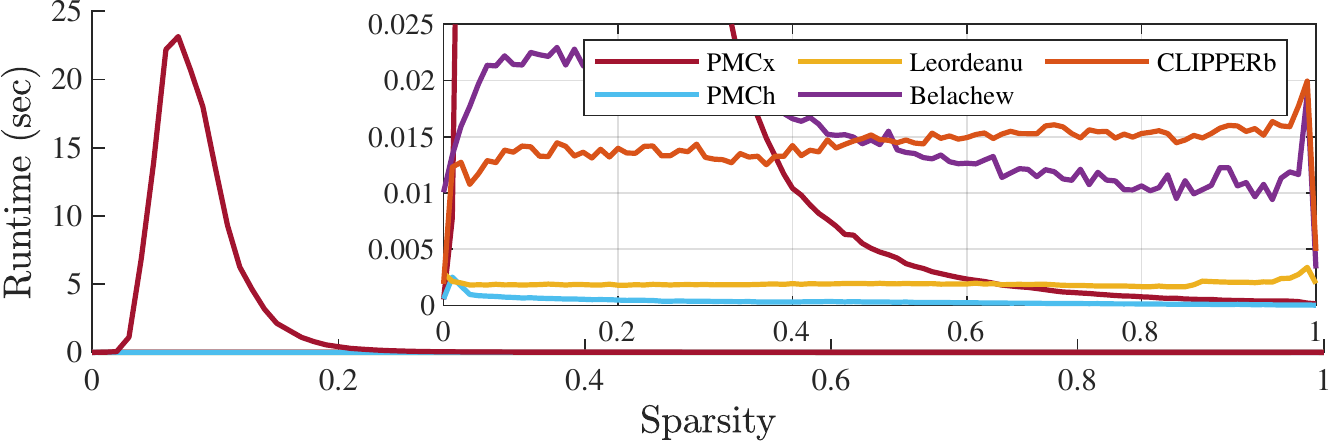}
\caption{
    Average algorithm runtime on binary graphs of size $n=200$ across $50$ Monte Carlo trials.
    When the input graph is not sparse, the exponential runtime of PMC Exact becomes clear.
    Compared to Belachew and PMC Exact, CLIPPER maintains a consistent runtime across all sparsity levels.
}
\vspace*{-0.3em}
\label{fig:runtime_vs_sparsity}
\end{figure}

\subsection{Maximum Clique Finding in Synthetic Data}
Given a data association problem and a scoring function $r(x)$ (see Fig.~\ref{fig:consistency_graph}c), factors such as noise and outlier statistics result in consistency graphs ranging from dense to sparse.
For example, in low outlier regimes, many associations are consistent and $\mathcal{G}$ is dense; conversely, in high outlier regimes with few consistent associations $\mathcal{G}$ is sparse.
In this section, we evaluate CLIPPER against the state of the art as an approximate MC solver on graphs of varying sparsity.
Graphs are generated from a complete graph that has increasingly more randomly selected edges removed according to the desired sparsity.
Evaluations of precision and runtime are given across $50$ Monte Carlo trials.
We emphasize that in this synthetic analysis, we are disregarding the ability of CLIPPER to find dense subgraphs in weighted graphs.

Fig.~\ref{fig:werr_vs_sparsity} shows the average clique size error of the approximate algorithms compared to the true clique size as provided by PMCx.
Positive error indicates overestimation.
In our evaluations, the graph size is limited to $n=200$ due to the high runtime of PMCx (see Fig.~\ref{fig:runtime_vs_sparsity}).
We observe that Leordeanu underestimates the clique size the most, while CLIPPERb and Belachew have the least average error across the entire sparsity range.
An algorithm's ability to find the largest clique (or densest subgraph) in a consistency graph directly affects its precision and recall in data association.

Fig.~\ref{fig:runtime_vs_sparsity} shows the corresponding runtime of each algorithm.
While PMCx (using 12 threads) is fast for sparser graphs, its runtime peaks in the low sparsity range due to the NP-hardness of exactly recovering the maximum clique.
As low graph sparsity corresponds to data association scenarios with few outliers, this runaway of runtime is problematic for a robot operating in varying conditions.
PMCh and Leordeanu acheive the fastest overall runtime, but as observed in Fig.~\ref{fig:werr_vs_sparsity}, underestimate the clique size the most on average.
CLIPPERb and Belachew strike a balance of precision and runtime across the sparsity range of binary graphs, with CLIPPERb performing consistently around \SI{15}{\milli\second}.

\begin{figure}[t!]
\centering
    \begin{subfigure}[b]{0.49\columnwidth}
        \centering
        \includegraphics[width=0.5\columnwidth]{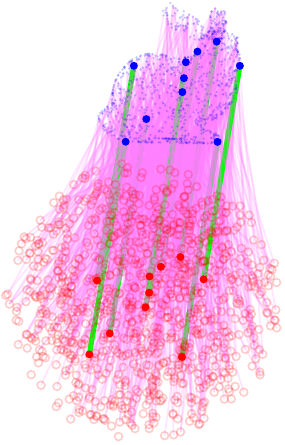}
        \label{fig:bunny:outliers}
    \end{subfigure}
    \begin{subfigure}[b]{0.49\columnwidth}
        \centering
        \includegraphics[width=0.5\columnwidth]{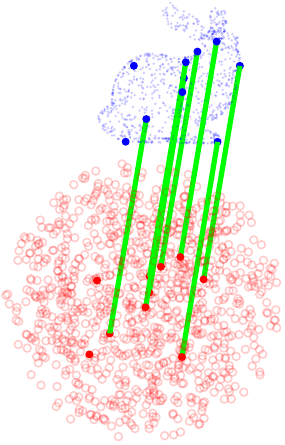}
        \label{fig:bunny:clippered}
    \end{subfigure}
    \caption{Robust data association using CLIPPER on Stanford Bunny.
    Only $10$ of $1000$ blue points in view 1 were seen in view 2 (discs), simulating a partial view with 1000 additional outlier points (circles).
    (left) Putative associations between the two point clouds having 99\% outlier associations, shown by magenta lines, and 1\% inlier associations, shown by green lines.
    (right) CLIPPER removes all outlier associations with 80\% recall in \SI{138}{\ms}.
    }
    \vspace*{-0.4em}
	\label{fig:bunny}
\end{figure}

\subsection{Data Association for Bunny Dataset}

Using the Stanford Bunny~\cite{curless1996stanfordbunny}, seen in Fig.~\ref{fig:bunny}, we evaluate CLIPPER as a data association algorithm in varying outlier regimes.
The Bunny model is first scaled to fit in a \SI{1}{\m} cube and $1000$ points are randomly sampled.
These points are arbitrarily rotated and translated into a second view, where noise uniformly sampled from $[\SI{-1}{\cm},\SI{1}{\cm}]$ is added.
For a model scale of \SI{1}{\m}, this level of noise could reasonably be expected from a sensor.
Additionally, 200 outlier points randomly drawn from a \SI{1}{\m} radius sphere are added in \hbox{view 2} to simulate clutter.
From the set of all-to-all associations, associations are randomly drawn according to the desired outlier ratio.
This process creates putative associations that could have been generated from feature matching or, in the case of no prior information, an all-to-all hypothesis.
Note that this is in stark contrast to ICP~\cite{besl1992method,chen1992object}, which performs poorly unless good initial information is provided.

To generate the consistency graph, pairwise consistency scores are computed as described in Section~\ref{sec:affinity-matrix}.
Weighted algorithms use the affinity matrix defined by ${s(x) \eqdef \exp(-\frac{1}{2}\frac{x^2}{\sigma^2})}$ for $|x|\le\epsilon$, and 0 otherwise.
Binary algorithms use the affinity matrix defined by $r(x)\eqdef1$ for $|x|\le\epsilon$, and 0 otherwise.
To study the effects of outliers, we choose $\epsilon=\SI{8}{\cm}$ and $\sigma=\SI{3}{\cm}$ small enough so that all algorithms have 100\% precision in the 0\% outlier regime.

\begin{figure}[t!]
\centering%
	\begin{subfigure}[b]{1\columnwidth}
	    \includegraphics[width=1\textwidth]{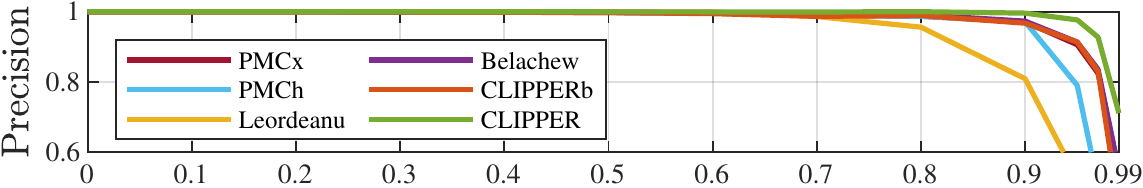}
	\end{subfigure}
	\begin{subfigure}[b]{1\columnwidth}
	    \includegraphics[width=1\textwidth]{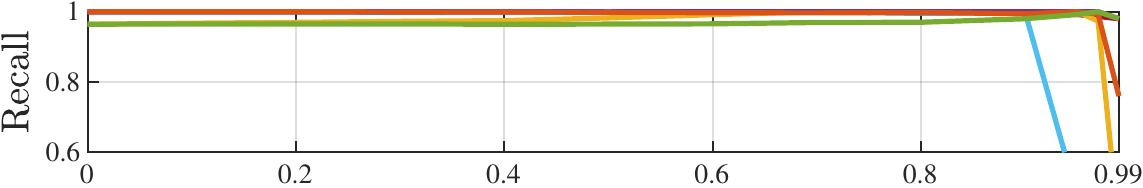}
	\end{subfigure}
	\begin{subfigure}[b]{1\columnwidth}
	    \includegraphics[width=1\textwidth]{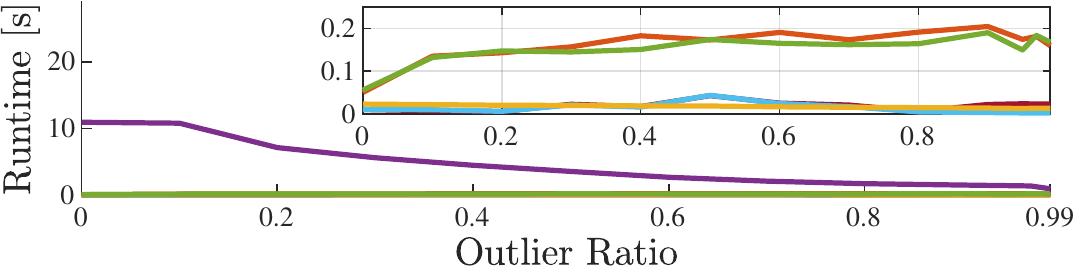}
	\end{subfigure}
\caption{Precision and recall for different outlier association ratios using $n=1000$ randomly drawn associations from the Stanford Bunny dataset.
}
	\label{fig:fullor_PR}
\end{figure}

By definition, precision $p\in[0,1]$ is the ratio of correct associations to the total number of associations returned by an algorithm, and recall $r\in[0,1]$ is the ratio of correct associations in an algorithm's output to the total number of associations in the ground truth.
The best performance is achieved when both precision and recall are high.
For many data association scenarios, precision is particularly important because a single outlier can have disastrous effects (e.g., loop closure in SLAM~\cite{mangelson2018pairwise}).
Precision, recall, and timing results vs outlier ratio are reported in Fig.~\ref{fig:fullor_PR}, with precision/recall results called out in Table~\ref{tbl:fullor_pr}.
CLIPPER maintains 100\% precision up to 90\% outliers, leading in precision up to 99\% outliers.
We observe that CLIPPER and Leordeanu have a recall of 97\% at 0\% outliers, due to their use of the weighted scoring function $s(x)$.
CLIPPER performs as expected in runtime, at about \SI{150}{\milli\second} on average.
Belachew, however, takes significantly longer, requiring \SI{10}{\second} at 0\% outliers and at least \SI{1}{\second} until a much sparser graph at 97\% outliers.

\begin{figure}[t!]
\centering%
    \includegraphics[width=1\columnwidth]{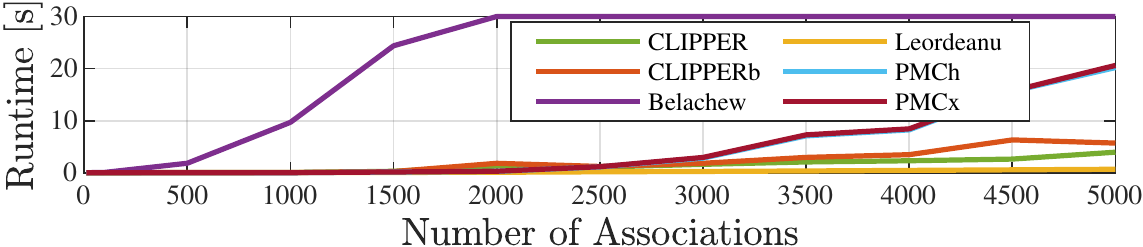}
\caption{Average runtime for increasing problem size with Bunny data in $0$--$10$\% outlier regime.
As the size of the input graph increases, CLIPPER runtime remains below \SI{4}{\s} while PMC Exact and Heuristic both take up to \SI{20}{\s}.
Belachew was terminated early at \SI{30}{\s} if not finished.}
	\label{fig:lowor-timing}
\end{figure}

A timing comparison is given in Fig.~\ref{fig:lowor-timing} with varying number of associations.
The Bunny dataset is used with a randomly sampled outlier ratio in $0$--$10$\% and is averaged across 5 Monte Carlo iterations.
While PMC quickly achieves the maximum clique for sparse problems (see Fig.~\ref{fig:runtime_vs_sparsity}), its runtime can increase without bound in large, dense problem domains.

\begin{figure}[t!]
\centering%
	\begin{subfigure}[b]{0.49\columnwidth}
	    \includegraphics[width=1\textwidth]{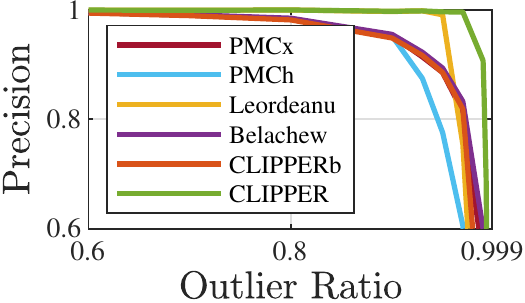}
	\end{subfigure}
	\begin{subfigure}[b]{0.49\columnwidth}
	    \includegraphics[width=1\textwidth]{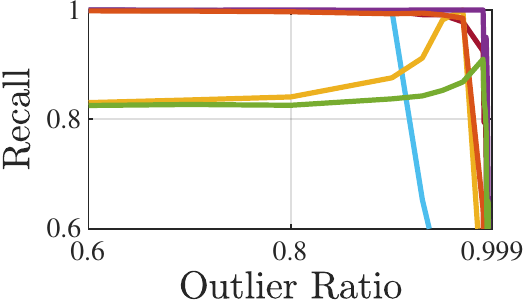}
	\end{subfigure}
\caption{Precision and recall of Bunny data association in high outlier regime.
By leveraging CLIPPER's weighted graph formulation, the affinity scoring function can be shaped to trade off recall for precision.}
\vspace*{-0.3em}
	\label{fig:highor}
\end{figure}

Fig.~\ref{fig:highor} illustrates the benefit gained by having a weighted affinity matrix, which is the flexibility to shape the scoring function $s(x)$ as opposed to the rigidity of using $r(x)$.
By varying the shape of $s(x)$, the user can trade off recall for precision in very high outlier regimes.
Fig.~\ref{fig:bunny} shows CLIPPER removing all outlier associations in this regime.

\begin{table}[!t] %
\scriptsize
\centering
\caption{
Precision and recall of CLIPPER against the state of the art for various outlier ratios (OR).
For each algorithm, results are reported as P/R.
The highest precision for each OR is shown in bold.
See Fig.~\ref{fig:fullor_PR}.
}
\setlength{\tabcolsep}{2.5pt}
\begin{tabular}{c c c c c c c c c c c c c c}
\toprule
OR  & CLIPPER && CLIPPERb && Belachew && Leordeanu && PMCh && PMCx \\ \toprule
   $0$  &  \textbf{1.00}/0.96  &&  \textbf{1.00}/1.00  &&  \textbf{1.00}/1.00  &&  \textbf{1.00}/0.96  &&  \textbf{1.00}/1.00  &&  \textbf{1.00}/1.00 \\
$0.70$  &  \textbf{1.00}/0.97  &&  0.99/1.00  &&  0.99/1.00  &&  0.99/1.00  &&  0.99/1.00  &&  0.99/1.00 \\
$0.80$  &  \textbf{1.00}/0.97  &&  0.99/1.00  &&  0.99/1.00  &&  0.96/1.00  &&  0.99/1.00  &&  0.99/1.00 \\
$0.90$  &  \textbf{1.00}/0.98  &&  0.97/0.99  &&  0.97/1.00  &&  0.81/1.00  &&  0.97/1.00  &&  0.97/1.00 \\
$0.95$  &  \textbf{0.98}/0.99  &&  0.91/1.00  &&  0.91/1.00  &&  0.53/1.00  &&  0.79/0.47  &&  0.91/0.99 \\
$0.97$  &  \textbf{0.93}/1.00  &&  0.83/0.99  &&  0.84/1.00  &&  0.32/0.97  &&  0.51/0.34  &&  0.82/0.99 \\
$0.99$  &  \textbf{0.71}/0.98  &&  0.44/0.76  &&  0.55/0.98  &&  0.04/0.38  &&  0.16/0.22  &&  0.55/0.98 \\
\bottomrule
\end{tabular}
\label{tbl:fullor_pr}
\end{table}

\section{CONCLUSION}\label{sec:conclusion}

We presented CLIPPER, a graph-theoretic framework for robust data association using the notion of geometric consistency.
CLIPPER was shown to consistently execute with low runtime and to outperform the state of the art, achieving 100\% precision, 80\% recall in 99\% outlier regimes.
These gains were found by implementing an efficient projected gradient descent algorithm and by formulating the data association problem on weighted graphs rather than binary.

\balance %

\bibliographystyle{IEEEtran}
\bibliography{refs}

\end{document}